\documentclass[journal,transmag]{IEEEtran}

\usepackage{amssymb}
\usepackage{amsthm,amsmath,amsfonts}
\usepackage{epsfig}
\usepackage{epstopdf}
\usepackage{chngcntr}

\counterwithin{table}{subsection}

\usepackage{amssymb}
\usepackage{subfigure}
\usepackage{etoolbox}
\usepackage{algorithmic}
\usepackage[algoruled,vlined]{algorithm2e}
\makeatletter
\renewcommand{\p@subfigure}{\thefigure}
\makeatother

\newcommand{\repeatable}[2]{\makeatletter \global\expandafter\def\csname repText@#1\endcsname {#2} \makeatother #2}
\newcommand{\repeatxt}[1]{\makeatletter \expandafter\csname repText@#1\endcsname \makeatother}

\newtoggle{citeInAbstract}
\toggletrue{citeInAbstract}

\newcommand{\usecrop}[2]
{
	\newlength{\cropwidth}
	\setlength{\cropwidth}{\the\textwidth}
	\addtolength{\cropwidth}{#1}
	\newlength{\cropheight}
	\setlength{\cropheight}{\the\textheight}
	\addtolength{\cropheight}{#2}
	\usepackage[width=\the\cropwidth,height=\the\cropheight,center]{crop}
}

\usepackage{totcount}

\usepackage{atbegshi}
\newif\ifRP
\newbox\RPbox
\setbox\RPbox\vbox{\vskip1pt}
\makeatletter
\AtBeginShipout{%
  \ifRP
    \AtBeginShipoutDiscard%
    \global\setbox\RPbox\vbox{\unvbox\RPbox
      \box\AtBeginShipoutBox\kern\c@page sp}%
  \fi
}%
\renewcommand{\RPtrue}{%
  \clearpage
  \ifRP\RPfalse\fi
  \global\let\ifRP\iftrue
}%
\renewcommand{\RPfalse}{%
  \clearpage
  \global\let\ifRP\iffalse
  \setbox\RPbox\vbox{\unvbox\RPbox
    \def\protect{\noexpand\protect\noexpand}%
    \@whilesw\ifdim0pt=\lastskip\fi
      {\c@page\lastkern\unkern\shipout\lastbox}%
  }%
}%
\makeatother

\DeclareMathAlphabet{\mathpzc}{OT1}{pzc}{m}{it}

\newcommand {\defeq}{\triangleq}

\newcommand {\myvec}[1] {{\mbox{\boldmath $#1$}}}
\newcommand {\mymat}[1]  {{\mbox{\boldmath $#1$}}}

\begin{document}

\title{Multi-View Kernels for Low-Dimensional Modeling of Seismic Events }

\author{Ofir~Lindenbaum,
        Yuri~Bregman, Neta~Rabin,
        and~Amir~Averbuch,~\IEEEmembership{Member,~IEEE}
\thanks{O. Lindenbaum was with the Department
of Electrical Engineering, Tel Aviv University, Israel,
P.O. Box 39040,
Tel-Aviv, 69978, Israel. e-mail: ofirlin@gmail.com.}
}

\markboth{}%
{}

\IEEEtitleabstractindextext{%
\begin{abstract}
The problem of learning from seismic recordings has been studied for years. There is a growing interest of developing automatic mechanisms for identifying the properties of a seismic event. One main motivation is the ability have a reliable identification of man-made explosions. The availability of multiple high dimensional observations has increased the use of machine learning techniques in a variety of fields. In this work, we propose to use a kernel-fusion based dimensionality reduction framework for generating meaningful seismic representations from raw data. The proposed method is tested on 2023 events that were recorded in Israel and in Jordan. The method achieves promising results in classification of event type as well as in estimating the location of the event. The proposed fusion and dimensionality reduction tools may be applied to other types of geophysical data.
\end{abstract}

}

\maketitle

\IEEEdisplaynontitleabstractindextext

\IEEEpeerreviewmaketitle

\section{Introduction}


\IEEEPARstart{M}{achine} learning techniques play a central role in data analysis, data fusion and visualization. As geophysical acquisition tools become more sophisticated and gather more information, data analysts relay more on machine learning techniques for
generating meaningful representations of the data. A coherent representation of complex data often includes a feature extraction step followed by a dimensionality reduction step, which results in a compact and visual model. 
Analysis tasks such as clustering, classification, anomaly detection or regression may be carried out in the constructed low-dimensional space. Common dimensionality reduction methods such as Principal Component Analysis (PCA) \cite{Pca1} and Linear Discriminant Analysis (LDA) \cite{LDA_book} project the feature space into a low dimensional space by constructing meaningful coordinated that are linear combinations of the original feature vectors. 
PCA is widely used for low-dimensional modeling of geoscience datasets. Jones \& Christopher \cite{Jones} applied PCA to infer aerosol specification for research of oceans or more complex land surfaces. Griparis and Faur \cite{Griparis} applied a linear dimensionality reduction tool,  Linear Discriminant Analysis (LDA) for a projection of earth observations into a low-dimensional space. Their low-dimensional representation resulted in a cluster organization of the image data by land types. PCA and Self organization maps \cite{Kohonen} were applied for pattern recognition in volcano seismic spectra by Unglert et. al. \cite{Unglert} and for geologic pattern recognition by Roden et. al. \cite{Roden}.

Another key issue in processing large amounts of data is the ability to fuse data from different sensors.  Typical seismometers record data using three channels. These three channels capture the motion in the horizontal and perpendicular directions to the earth. Each channel may be processed separately and the results can be combined. Alternatively, a fused representation may be formed for common analysis. Recent advances in machine learning and in particular the use of non-linear kernel-based algorithm enable to construct data-driven fusions and to compute geometry-preserving low-dimensional embeddings. Such kernel-based embedding techniques are known as manifold learning methods, among them Local Linear Embedding \cite{LLE}, Lapacian Eigenmaps \cite{Belkin1} and Diffusion Maps (DM) \cite{Lafon}. Manifold learning methods overcome limitations of linear dimensionality reduction tools such as PCA and LDA \cite{manifold_book}. When the relationship between the original high-dimensional points is complex and non-linear, linear projections may fail to organize the data in a way that is loyal to the intrinsic physical parameters that drives the observed phenomena.

This work focuses on extending manifold learning techniques for low-dimensional modeling and kernel based data-driven fusion of seismic data. Identifying the characteristic of seismic events is a challenging and important task. This includes the discrimination between earthquakes and explosions which is not only an essential component of nuclear test monitoring but it is also important for the maintaining the quality of earthquake catalogs. For example, wrong classification of explosions as earthquakes may cause the erroneous estimation of seismicity hazard. The discrimination task is typically performed based on some extracted seismic parameters. Among such parameters is the focal depth, the ratio between surface wave magnitude and body wave magnitude and the spectral ratio between different seismic phases \cite{Blandford}, \cite{rodgers1997comparison}.
Discrimination methods based on seismic parameters give only a partial solution to the problem. For instance, a larger half of seismic events reported by the Comprehensive Nuclear-Test-Ban Treaty Organization (CTBTO) are not “screened out” as natural events or even are not considered for the discrimination at all although most of those events are typically earthquakes \cite{BenHorin}.

Recently, this  problem and other geophysical challenges have been approached using machine
learning frameworks. Hidden Markov model were proposed in \cite{Ohrnberger}, \cite{Beyreuther}, \cite{Hammer} and modeled the data in an unsupervised manner.  Artificial neural networks \cite{Tiira}, \cite{DelPezzo}, \cite{esposito2006automatic} or support vector machines \cite{Kortstrom,ruano2014seismic} were also used to construct a classifier in a supervised manner. The study in \cite{Kuyuk} utilizes Self Organization Maps to distinguish micro-earthquakes from quarry blasts in the vicinity of Istanbul, Turkey. Manifold learning is used in \cite{ramirez2011machine} for seismic phase classification.  In \cite{mishne2015graph} a graph is used to detect sea mines in side-scan sonar images. The DM method is used in \cite{fernandez2015diffusion} for visualization of meteorological data. A non-linear dimensionality reduction is proposed in \cite{rabin2016earthquake} to discriminate between earthquakes and explosions.

In this study, the manifold learning approach that was presented in \cite{rabin2016earthquake} is extended by using a kernel-based fusion method for identification of seismic events. The method is model-free, and it is based on signal processing for feature extraction followed by manifold learning techniques for embedding the data. Furthermore, the method reviles the underlying intrinsic physical properties of the data, which results in a natural organization of the events by type. Since seismic data is recorded at multiple channels, we suggest fusing the information to extract a more reliable representation for the seismic recordings. The fusion framework is based on a recent work by \cite{lindenbaum,lindenbaum2015learning}. The study extends Diffusion Maps (DM) \cite{Lafon}, which has been successfully applied for phase classification \cite{ramirez2011machine}, for estimation of arrival times \cite{taylor2011estimation} and for events discrimination \cite{Poster}. Other constructions for fusing kernels were proposed in \cite{salhov2016multi,lederman,michaeli}.

The proposed framework begins with a preprocessing stage in which a time-frequency representation is extracted from each seismic event. The training phase includes the construction of a normalized graph that holds the local connections between the seismic events. A low dimensional map is then obtained by the eigen-decomposition of the graph. The constructed embedding is distance preserving. Thus the geometry of the dataset is kept in the new embedding coordinates. By utilizing the low dimensional embedding, we demonstrate capabilities of classification, location estimation and anomaly detection of seismic events.  

The paper is organized as follows: Sections \ref{sec:Manifold} and \ref{SecMulti} present the machine learning frameworks for manifold learning and data fusion. In Section \ref{sec:Data} the data set is described. The mathematical methods required for analysis of seismic data are provided in Section \ref{sec:Pre}. The proposed framework and experimental results are presented in Section \ref{sec:Exp}. We conclude this work in Section \ref{sec:Future}.

\section{Manifold Learning }
\label{sec:Manifold}

This section reviews the manifold learning method that is applied in this work for non-linear dimensionality reduction, diffusion maps. The method's main ingredient is a kernel function. Here, radial basis kernel functions are used, their construction is described in detail. 

\subsection{Radial Basis Kernel Function}
Kernel functions are vastly utilized in machine learning. Classification, clustering and manifold learning use some affinity measure to learn the relations among data points. A kernel is a pre-defined similarity function designed to capture the fundamental structure of a high dimensional data set.
Given a high dimensional data set \begin{math} {\myvec{X} =  \lbrace
	{ { \myvec{x}_1,\myvec{x}_2, \myvec{x}_3,...,\myvec{x}_M  }} \rbrace
},\myvec{x}_i	\in {\mathbb{R}^{D}},
\end{math} a kernel ${{\cal{K}} : \mymat{X}\times{\mymat{X}}\longrightarrow{\mathbb{R}}  }$ is an affinity function over all pairs of points in $\mymat{X}$. The discrete kernel is represented by a matrix $\mymat{K}$ with following properties

\begin{itemize} 

\item {Symmetry \begin{math}{K_{i,j}={\cal{K}}(\myvec{x}_i,\myvec{x}_j)={\cal{K}}(\myvec{x}_j,\myvec{x}_i) }
\end{math}} 
\item {Positive semi-definiteness: \begin{math}{ \myvec{v}_i^T  \mymat{K}  \myvec{v}_i \geq 0 }\end{math} for all $\myvec{v}_i \in
\mathbb{R}^M$ and \begin{math}{{\cal{K}}(\myvec{x}_i,\myvec{x}_j)
	\geq 0. }
\end{math}}

        \end{itemize}
These properties guarantee that the matrix $\mymat{K}$ has
		real eigenvectors and non-negative real eigenvalues.
In this study  radial basis functions (RBF) are used for constructing the kernel. The RBF kernel function is defined by
\begin{equation}\label{eq:RBF}
{ K_{i,j}=exp\{
	{-\frac{||\myvec{x}_i-\myvec{x}_j||^2}{2 \sigma^2}\}}  }.\end{equation}  Applying the Euclidean distance to high dimensional pairs of distant vectors could somewhat be misleading, as data is typically sparse in the high-dimensional space. For this reason the decaying property of the Gaussian kernel is beneficial. The Gaussian tends to zero for distant points, whereas its value is close to one for adjacent points.
\subsection{Setting the Kernel's Bandwidth}
The kernel's bandwidth $\sigma$ controls the number of points taken into consideration by the kernel. 
A simple choice for \begin{math} \sigma \end{math} is based on the standard deviation of the data. This approach is good when the data is sampled from a uniform distribution. In this study, we use a max-min measure. The method was proposed in \cite{Keller} and aims to find a small scale to maintain local connectivities. The scale is set to
\begin{equation} \label{eq:MaxMin}
\sigma^2_{\text{MaxMin}}={\cal{C}}\cdot \underset{j}{\max} [ \underset{i,i\neq j}{\min} (||\myvec{x}_i-\myvec{x}_j||^2)],
\end{equation}
where ${\cal{C}} \in [2,3]$. Alternative methods such as \cite{Singer,lindenbaum2015musical} have demonstrated similar results in our experiments.

\subsection{Non-Linear Dimensionality Reduction}
\label{SecDiff}
Most dimensionality reduction methods are unsupervised frameworks that seek for a low dimensional representation of complex, high dimensional data sets. Each method preserves a certain criteria while reducing the dimension of the data. Principal component analysis (PCA) \cite{PCA},  reduces the dimension of the data while preserving most of the variance. Non linear methods such as Local Linear Embedding \cite{LLE}, Laplacian Eigenmaps
\cite{Luo}, Diffusion Maps (DM) \cite{Lafon} preserve the local structure of the high-dimensional data. In particular, in DM \cite{Lafon}, a metric that describes the intrinsic connectivity between the data points is defined. This metric is preserved in the low-dimensional space, resulting in a distance-preserving embedding. The metric is refereed to as diffusion distance, it is defined later in this subsection.

The DM framework enforces a fictitious random walk on the graph of a high dimensional data set $\myvec{X}=\{\myvec{x}_1,..,\myvec{x}_M \},\myvec{x}_i \in \mathbb{R}^D$. This results in a Markovian process that travels in the high-dimensional space only in areas where the sampled data exists. The method has been demonstrated useful when applied to audio signals \cite{lindenbaum2015musical}, image editing \cite{farbman2010diffusion}, medical data analysis \cite{haghverdi2015diffusion} and other types of data sets.

Reducing the dimension of a data set by construction of DM coordinates is performed using the following steps
\begin{enumerate}
\item Given a data set $\myvec{X}$ compute an RBF kernel $\myvec{K}$ based on Eq. \ref{eq:RBF}.
\item Normalize the kernel using  $\mymat{D}$ where  \begin{math} D_{i,i}=\underset{j}{\sum}{K_{ij}} \end{math}. 
Construct the row stochastic matrix $\mymat{P}$ by \begin{equation}
{P_{i,j}\defeq{\cal{P}}(\myvec{x}_i,\myvec{x}_j)\defeq[{{\mymat{D}}^{-1}{\mymat{K}}  }}]_{i,j}
\label{EquationPDM}
.\end{equation} 
\item  Compute the spectral decomposition of the matrix \begin{math}  \mymat{P} \end{math} to obtain a sequence of eigenvalues  \begin{math}{\lbrace {\lambda_m}\rbrace }
\end{math} and normalized right eigenvectors \begin{math}{\lbrace{{\mbox{\boldmath${\psi}$}}_m}\rbrace }
\end{math} that satisfy ${ {\mymat{P}}  {\mbox{\boldmath${\psi}$}_m} =\lambda_m{\mbox{\boldmath${\psi}$}}_m, m=0,...,M-1}
$;
\item Define the $d$-dimensional ($d \ll D$) DM representation as
\begin{equation}\label{EQPSI}{ \myvec{\Psi}{(\myvec{x}_i)}:   \myvec{x}_i
	\longmapsto \begin{bmatrix} { \lambda_1\psi_1(i)} , {.} {.} {.}
	,
	
	{\lambda_{d}\psi_{d}(i)}\\
	
	\end{bmatrix}^T \in{\mathbb{R}^{d}} },
\end{equation}
where $\psi_m(i)$ denotes the $i^{\rm{th}}$ element of ${\mbox{\boldmath${\psi}$}_m}$.
\end{enumerate}
The power of the DM framework stems from the Diffusion Distance (Eq. \ref{EqDist}). It was shown in \cite{Lafon} that the Euclidean distance in the embedded space $ \myvec{\Psi}{(\myvec{x}_i)}$ is equal to a weighted distance between rows of the probability matrix $\myvec{P}$. 
	This distance is defined as the Diffusion Distance
	\begin{equation}{ \label{EqDist} { {\cal{D}}^2_t( \myvec{x}_i,\myvec{x}_j)=||{\mymat{\Psi}_t{(\myvec{x}_i)}}-{\mymat{\Psi}_t{(\myvec{x}_j)}}||^2=||\myvec{P}_{i,:}-\myvec{P}_{j,:}||^2_{\tiny\mymat{W}^{-1}}},}
			\end{equation}
			where $\mymat{W}$ is a diagonal matrix with elements
			$W_{i,i}=\frac{D_{i,i}}{\sum_{i=1}^M D_{i,i}}$. Thus, the DM embedding is distance preserving, meaning that neighboring points in the high-dimensional space are embedded close to each other in the diffusion coordinates.

\section{Data Fusion}
\label{SecMulti}
Many physical phenomena are sampled using multiple types of sensing devices. Each sensor provides a noisy measurement of a latent parameter of interest. Data fusion is the process of incorporating multiple observation of the same data points to find a more coherent and accurate representation. \\
{\bf{Problem Formulation:}} Given multiple sets of data points  $\mymat{X}^l\text{ }, l=1,...,L$. Each
view is a high dimensional dataset ${\mymat{X}^l = \lbrace{ {
			\myvec{x}_1^l,\myvec{x}_2^l, \myvec{x}_3^l,...,\myvec{x}_M^l  }}
	\rbrace , \myvec{x}_i^l\in {\mathbb{R}^{D}} }$. Find a reliable low dimensional representation $\myvec{\Psi}(\myvec{X}^1,...,\myvec{X}^L)\in \mathbb{R}^d$.

\subsection{Multi-View Diffusion Maps (Multi-View DM)}
\label{Sec2} An approach for fusion kernel matrices in the spirit of DM framework was presented in \cite{lindenbaum}. The idea is to enforce a random walk model based on the kernels that model each view by restraining
the random walker to ``hop'' between views in each time step.

The construction requires to compute a Gaussian kernel for each view
\begin{equation} \label{EQK}{ K^l_{i,j}=exp\{
	{-\frac{||\myvec{x}^l_i-\myvec{x}^l_j||^2}{2 \sigma_l^2}\}},\text{ } l=1,...,L },\end{equation}
then the multi-view kernel is formed by the following matrix
\begin{equation} \label{EQKMAT}
\mymat{\widehat{K}}= \begin{bmatrix}  \mymat{0}_{M \times M} & {\mymat{K}^1\mymat{K}^2}&  {\mymat{K}^1\mymat{K}^3}&...& {\mymat{K}^1\mymat{K}^p} \\
\mymat{K}^2\mymat{K}^1 & \mymat{0}_{M \times M} & {\mymat{K}^2\mymat{K}^3}&...& {\mymat{K}^2\mymat{K}^p}\\ \mymat{K}^3\mymat{K}^1 & {\mymat{K}^3\mymat{K}^2} &
\mymat{0}_{M \times M} &...& {\mymat{K}^3\mymat{K}^p}\\:&:&:&...&:\\\mymat{K}^p\mymat{K}^1 & {\mymat{K}^p\mymat{K}^2} & {\mymat{K}^p\mymat{K}^3} &...&
{\mymat{0}_{M \times M}}.
\end{bmatrix}. \end{equation}
Next, re-normalizing using the diagonal matrix $\mymat{\widehat{D}}$ where
\begin{math}
{\widehat{D}}_{i,i}=\underset{j}{\sum}{{\widehat{K}}_{i,j}}
\end{math}, the normalized row-stochastic matrix is defined as
\begin{equation}
\label{phat}
\mymat{\widehat{P}}={\mymat{\widehat{D}}}^{-1}\mymat{\widehat{K}}, ~~~
{\widehat{P}}_{i,j}=\frac{{{\widehat{K}}_{i,j}} }{\widehat{D}_{i,i} },
\end{equation}
where the $m,l$ block is a square $M\times M$ matrix located at\\
$[1+(m-1)M,1+(l-1)M], l=1,...,L$. This block describes the probability of
transition between view $\mymat{X}^m$ and $\mymat{X}^l$.
The multi-view DM representation for $\mymat{X}^l$ is computed
by
\begin{equation}
\label{Map1}
{ \myvec{\widehat{\Psi}}_t{(\myvec{x}^l_i)}:   \myvec{x}^l_i
	\longmapsto
	\begin{bmatrix} { \lambda_1^{t}\psi_1(i+\bar{l})} , {.} {.} {.} ,
	
	{\lambda_{d}^{t}\psi_{d}(i+\bar{l})}
	
	\end{bmatrix}^T \in{\mathbb{R}^{d}} },
\end{equation} where $\bar{l}=(l-1)\cdot M$.
The final low dimensional representation is defined by a concatenation of all low dimensional multi-view mappings
\begin{equation} \label{eq:MVDMrep} \vec{\myvec{\Psi}}(\vec{\myvec{X}})=
{[ \myvec{\widehat{\Psi}}{(\myvec{X}^1)}},\myvec{\widehat{\Psi}}{(\myvec{X}^2)},...,\myvec{\widehat{\Psi}}{(\myvec{X}^L)}].
\end{equation}

\subsection{Alternative Methods}
\label{SECALT}
Here we provide a brief description of several methods for fusing the views before the application of a spectral decomposition. \\
{\bf{Kernel Product (KP)}}: Multiplying the kernel matrices element wise  $\mymat{K}^{{\circ}}\defeq \mymat{K}^1
\circ \mymat{K}^2 \circ ...\circ \mymat{K}^L $, ${K}_{ij}^{\circ}\defeq {K}_{ij}^1
\cdot {K}_{ij}^2\cdot...\cdot {K}_{ij}^L$, then normalizing by the sum of rows. The resulting row stochastic matrix is denoted as $\mymat{P}^{{\circ}}$.
This kernel corresponds to the approach in \cite{Lafon}.\\
{\bf{Kernel Sum (KS)}}: Defining the sum
kernel $\mymat{K}^{+}\defeq\sum^L_{l=1}\mymat{K}^l$. Normalizing the sum kernel by the sum
of rows, to compute $\mymat{P}^{+}$. This random walk sums the step probabilities from each view.
This approach is proposed in \cite{Zhou}.\\
{\bf{Kernel Canonical Correlation Analysis (KCCA)}}: This method detailed in \cite{lai2000kernel,akaho2006kernel} extend the well know Canonical Correlation Analysis (CCA).Two kernels $\myvec{K}^1 \text{and }\myvec{K}^2$ are constructed in each view as in Eq. (\ref{EQK}) and the canonical vectors $\myvec{v}_1 \text{and }\myvec{v}_2$ are computed by solving the following generalized eigenvalue problem
		\begin{equation} \label{eq:KCCA}
\begin{bmatrix}  \mymat{0}_{M \times M} & {\mymat{K}^1\cdot \mymat{K}^2} \\ {{\mymat{K}^2\cdot \mymat{K}^1}} & \mymat{0}_{M \times M} \end{bmatrix} \begin{pmatrix}
{\myvec{v}_1} \\ {\myvec{v}_2}
\end{pmatrix}= \rho \cdot \begin{bmatrix}  (\mymat{K}^1+\gamma\myvec{I})^2 & {\mymat{0}_{M \times M}} \\ {\mymat{0}_{M \times M}} &(\mymat{K}^2+\gamma\myvec{I})^2  \end{bmatrix} \begin{pmatrix}
{\myvec{v}_1} \\ {\myvec{v}_2}
\end{pmatrix},
		\end{equation} where $\gamma \myvec{I}$ are regularization terms which guarantee that the matrices $(\mymat{K}^1+\gamma\myvec{I})^2$ and  $(\mymat{K}^2+\gamma\myvec{I})^2 $ are invertible.


\section{Seismic Data Set}
\label{sec:Data}

\begin{figure}
	\centering
	\includegraphics[scale = 0.45]{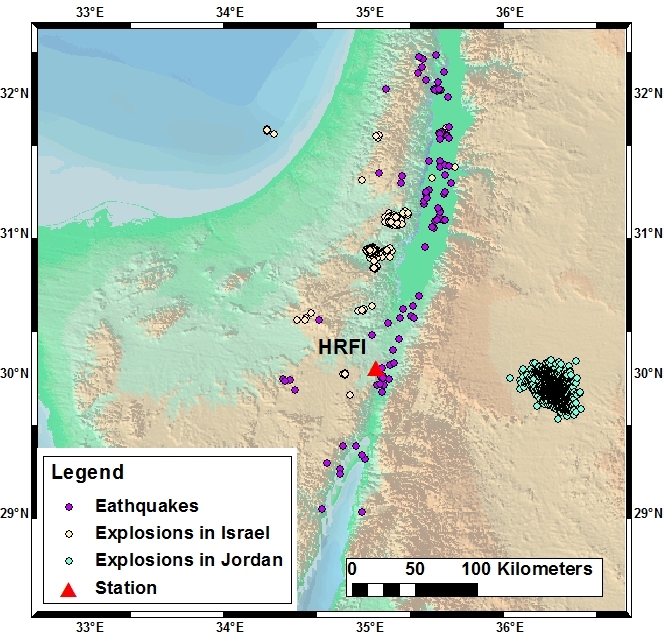}
	\caption{Seismic events in the data set and the HRFI station.}\label{fig:QuarryLocationsA}
\end{figure}

The data set that is used for demonstrating the proposed kernel based approaches includes $2023$ explosions and $105$ earthquakes. $1654$ of the explosions  occurred at the Shidiya phosphate quarry in the Southern Jordan between the years $2005$-$2015$ (see a map of the region in Figure \ref{fig:QuarryLocationsA}). These events were reported by the Israel National Data Center at the Soreq Nuclear Research Center with magnitudes $2\leq \text{ML}\leq3$ seismic. The rest of the events were taken from the seismic catalog of the Geophysical Institute of Israel between the years $2004$-$2014$ . All events were reported in Israel between latitudes $29^{\circ}$N-$32.5^{\circ}$N and longitudes $34.2^{\circ}$E-$35.7^{\circ}$E with duration magnitudes Md $\ge 2.5$. 

Most of the earthquakes in the dataset occurred in the Dead Sea transform \cite{Garfunkel}.
The dataset includes the February 11, 2004 earthquake with the duration magnitude of Md = 5.1.  This was the strongest event in this area since 1927 \cite{Hofstetter}. Twelve aftershocks that are included in the dataset are associated with this main shock. 
The majority of the explosions in the dataset are ripple-fire query blasts. Moreover, the dataset consists of several one shot explosions, for instance, two experimental underwater explosions in the Dead Sea \cite{Hofstetter} and surface and near-surface experimental explosions at the Oron quarry \cite{gitterman2009source} and at the Sayarim Military Range \cite{Fee}  in the Negev desert.

The dataset consists of  seismogram recordings from the HRFI (Harif) station in Israel. The station is part of the Israel National Seismic Network \cite{Hofstetter}. It is equipped with a three component broad band STS-2 seismometer and a Quanterra data logger. The seismograms are sampled at a frequency of $40$ Hz. Waveform segments of $2.5$-minutes  (6000 samples) have been selected for every event.  In each waveform, the first P phase onsets reside $30$ seconds after the beginning of each waveform.
Figure \ref{fig:QuarryLocationsA} displays the events on the regional map.

\section{Seismic Preprocessing and Feature Extraction Methods}
\label{sec:Pre}
This section provides background on typical methods that are used for seismic signal processing as well as the description of the feature extraction method that was applied here. First, the STA/LTA detector is reviewed. Next, we describe how the alignment between the different waveforms was implemented. Last, the feature extraction step, which results in a time-frequency representation of the seismic signal, is described.

\subsection{Short and Long Time Average (STA/LTA)}
\label{STA}
Detection of seismic signal embedded in the background noise is a classical problem in the signal processing theory. In the context of statistical decision theory it may be formulated as a choice between two alternatives: a waveform contains solely the noise or it contains a signal of interest superimposed on the noise.  The STA/LTA trigger is a most widely accepted detection algorithm in seismology \cite{trnkoczytopic}. It relies on the assumption that a signal is characterize by a concentration of higher energy level compared with the energy level of the noise. This is done by comparing short-time energy average to a long-time energy average using a Short Time Average/ Long Time Average (STA/LTA) detector. Usually a band-pass filter is applied before the STA/LTA test.

Given a time signal $\myvec{y}(n)$ the ratio $R(i)$ is computed at each time instance $i$ is computed as follows
\begin{equation} \label{eq:STALTA}
R(i)=\frac{L\cdot[\overset{i+S}{\underset{{j=i}}{\sum}}y^2(j)]}{S\cdot(\overset{i+L}{\underset{{j=i}}{\sum}}y^2(j))},
\end{equation} where $L \gg S$ are the number of samples used for the long and short average correspondingly. The ratio $R(i)$ is compared to a threshold $\delta$ to identify time windows suspected as seismic events.

\subsection{Seismic Event Alignment}
All waveform segments in the dataset were extracted according to the first P phase onset time. Those onset times were manually picked by the analysts. However, our selective waveform inspection showed that the P onsets often have actual offsets of several seconds, sometimes even of ten seconds. 
In order to increase the accuracy of the alignment,  Algorithm \ref{alg:Trigger} is proposed to detect the first P onsets.
\begin{algorithm}[h]
	\caption{Seismic trigger alignment} \textbf{Input:} Input time signals $\myvec{y}[n]$.\\ \textbf{Output:} Estimated time sample $\hat{n}_P$ for P onset of seismic event.
	\begin{algorithmic}[1]
		\STATE Apply a finite impulse response band pass filter to $\myvec{y}[n]$. The filter $\myvec{h}_1$ is designed to pass the signal between $f^{(1)}_L=2[Hz] \text{ and } f^{(1)}_H=4[Hz]$. The filtered signal is denoted as $\myvec{\tilde{y}}^{(1)}[n]$
        \STATE Compute the STA/LTA ratio based on Eq. (\ref{eq:STALTA}).
		\STATE Set $n^{(1)} \defeq \min  (n),\text{s.t. } R(n)> \delta$. The threshold $\delta$ is computed based on the following formula $\delta= \min (4,0.3 \cdot \max (R(n)))$.
        \STATE Repeat steps 1-3 using $f^{(2)}_L=4[Hz],f^{(2)}_H=8[Hz]$ and $f^{(3)}_L=8[Hz],f^{(3)}_H=12[Hz]$. Denote the trigger indexes as $n^{(2)}$ and $n^{(3)}$.
        \STATE Set the estimated trigger as $\hat{n}\defeq \min (n^{(1)},n^{(2)},n^{(3)})$.
		
	\end{algorithmic}
	\label{alg:Trigger}
\end{algorithm}

\begin{figure}
	\centering
	\includegraphics[scale = 0.2]{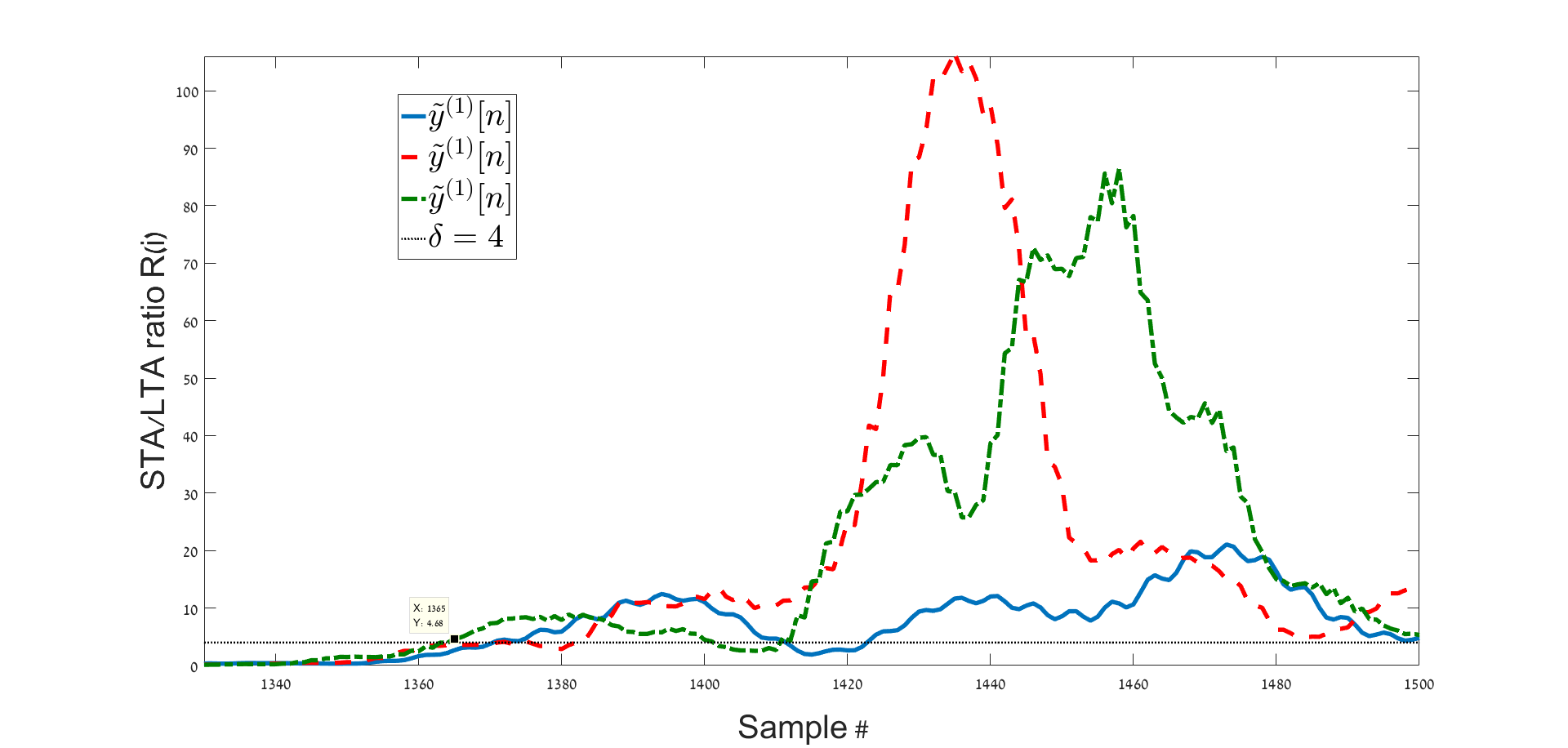}

	\caption{The STA/LTA ratios (Eq. \ref{eq:STALTA}) computed for an earthquake. Each ratio $R(i)$ is computed using one of three filtered signals $\myvec{\tilde{y}}^{(1)}(n),\myvec{\tilde{y}}^{(2)}(n),\myvec{\tilde{y}}^{(3)}(n)$. The filters are designed as explained in step 2 of Algorithm \ref{alg:Trigger}. The constant black line is an example of a threshold $\delta=4$. The onset $\hat{n}_P$ is defined as the first cross point of the threshold $\delta$.}\label{fig:STALTA}
\end{figure}

Algorithm \ref{alg:Trigger} aligns the seismic events based on the STA/LTA ratios which are computed using three filtered versions of the input signal. We assume that most of the energy of the seismic signature is between $2[Hz]$ and $12[Hz]$. Figure \ref{fig:STALTA} presents a visual example for the application of Algorithm \ref{alg:Trigger}.

\subsection{Feature Extraction}
\begin{figure}
	\centering
	\includegraphics[scale = 0.25]{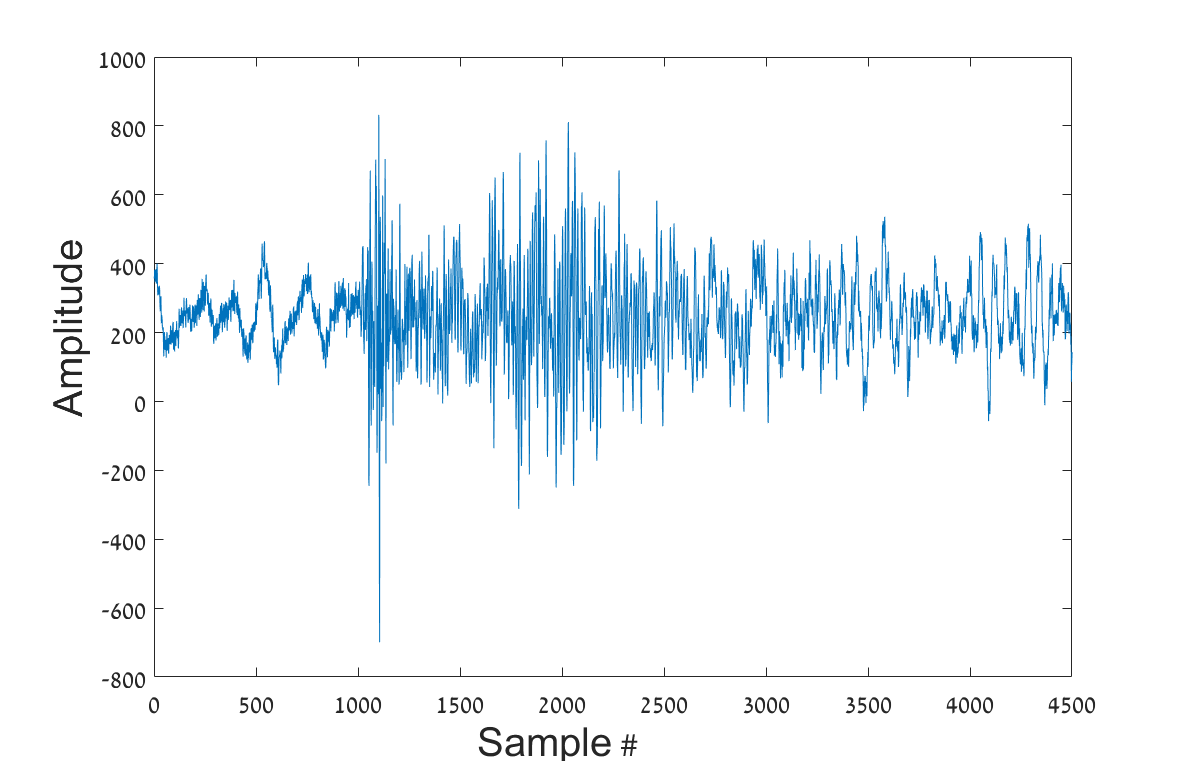}
	\includegraphics[scale = 0.25]{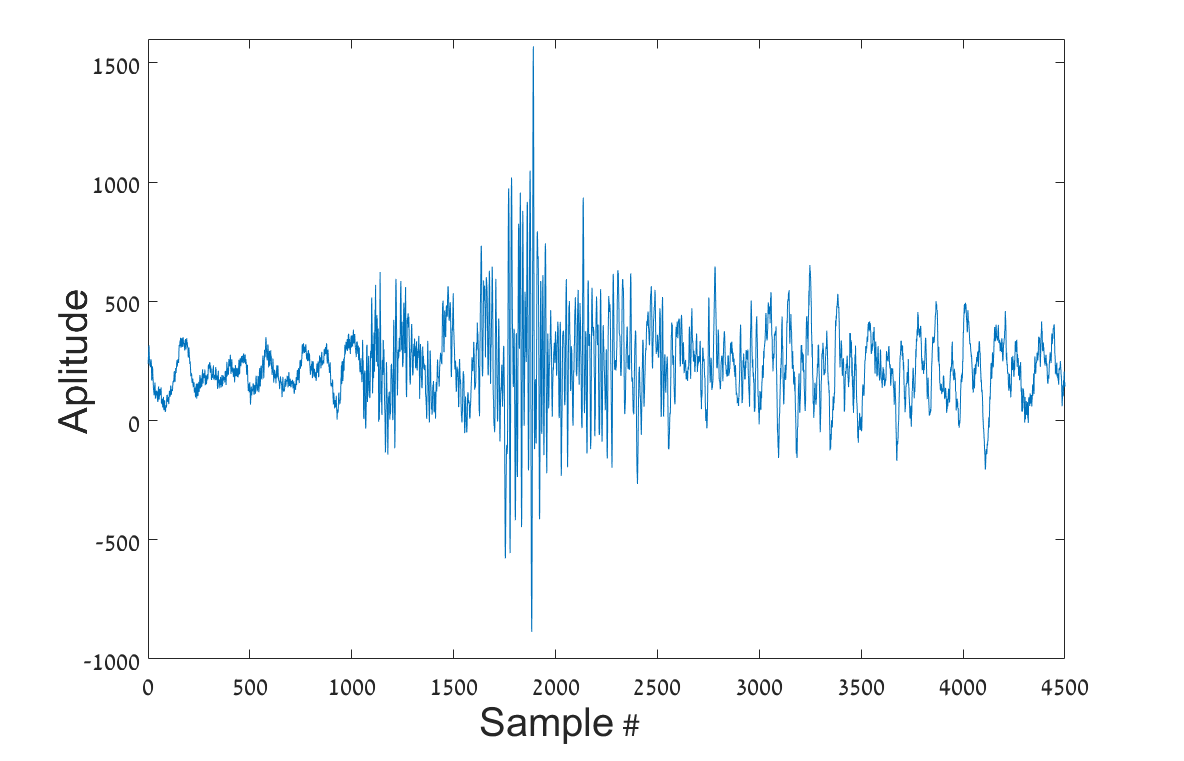}
	\includegraphics[scale = 0.25]{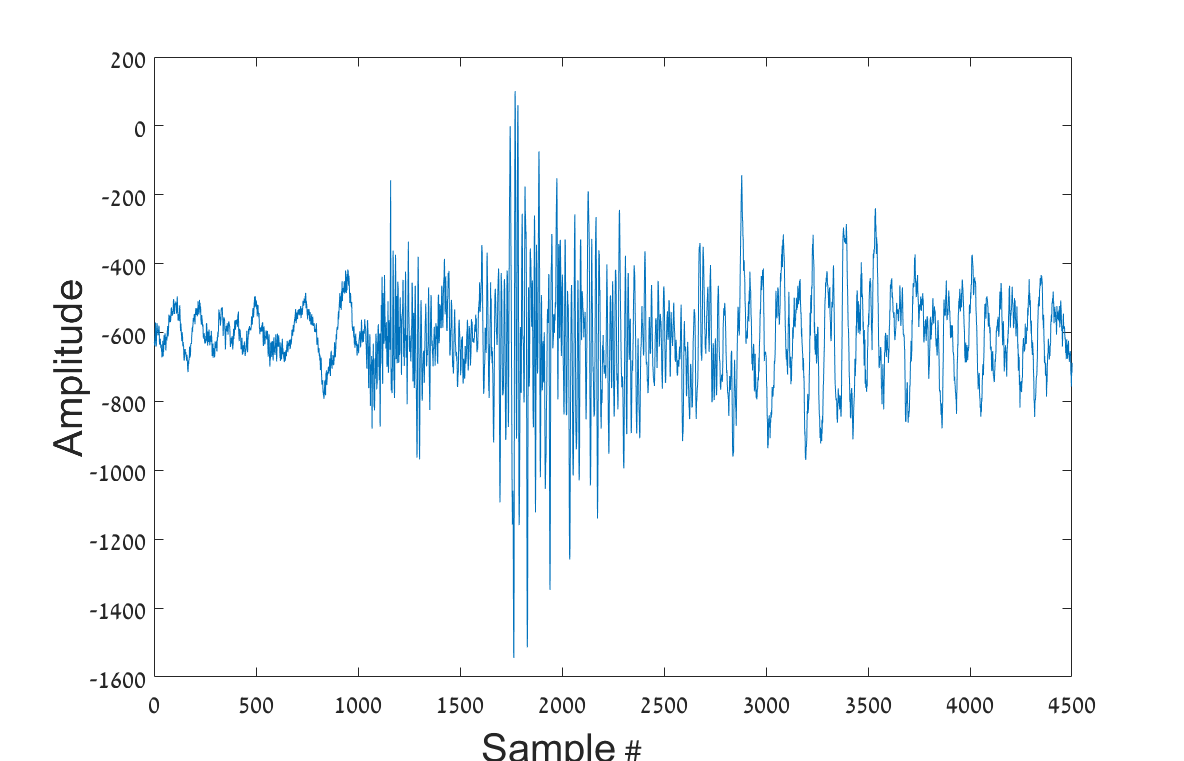}
	\caption{4500 samples from a recording of an explosion. Top - Z channel. Middle - E channel. Bottom - N channel.}\label{fig:Sig}
\end{figure}

In this study a time-frequency representations, named sonograms \cite{joswig1990pattern}, is used, with some modification. The sonogram is a normalized short time Fourier transform (STFT) rearranged to be equal tempered on a logarithmic scale. 
Each raw single-trace seismic waveform input is denoted by $\myvec{y}(n)\in \mathbb{R}^{\bar{N}}$. The length of the signals in this study is $N=6,000$ with a sampling rate of $F_s=40 Hz$. An example of seismic signals recorded using three channels is presented in Figure \ref{fig:Sig}.

The sonogram is extracted from $\myvec{y}(n)$ based on the following steps:
\begin{itemize}

\item {Given a recorded signal $\myvec{y}(n)\in \mathbb{R}^N$ the short time Fourier transform (STFT) is computed by
\begin{equation}
\label{eq:STFT}
\myvec{{{STFT}}}(f,t) = \sum\limits_{n = 1}^{N} {w(n-\ell) \cdot {y}(n)}  \cdot {e^{ - j2\pi f}} ,
\end{equation}
where $w(n-t)$ is a Hann window function with a length of $N_0=256$ and a $s=0.8$ overlap. The time indexes are $\ell=(1-s)\cdot N_O\cdot t ,t=1,...,T$. The number of time bins is computed using the following equation
\begin{equation}
\label{eq:TimeWin}
T=\lceil{\frac{N-N_0}{(1-s)\cdot N_0}}\rceil+1
\end{equation}}
\item{The Spectrogram is the normalized energy of $\myvec{STFT}(f,t)$
\begin{equation}
\label{eq:NSPEC}
\myvec{R}(f,t) = \frac{\myvec{STFT}(f,t)^2 }{N_0}.
\end{equation} The Spectrogram $\myvec{R}(f,t)$ contains $T$ time bins and $F=N_0$ frequency bins. }
\item{The frequency scale is then rearrange to be equally tempered on a logarithmic scale, such that the final spectrogram contains $11$ frequency bands. The frequency bands are presented in Table \ref{table0}.

\begin{figure}
	\centering
	\includegraphics[scale = 0.26]{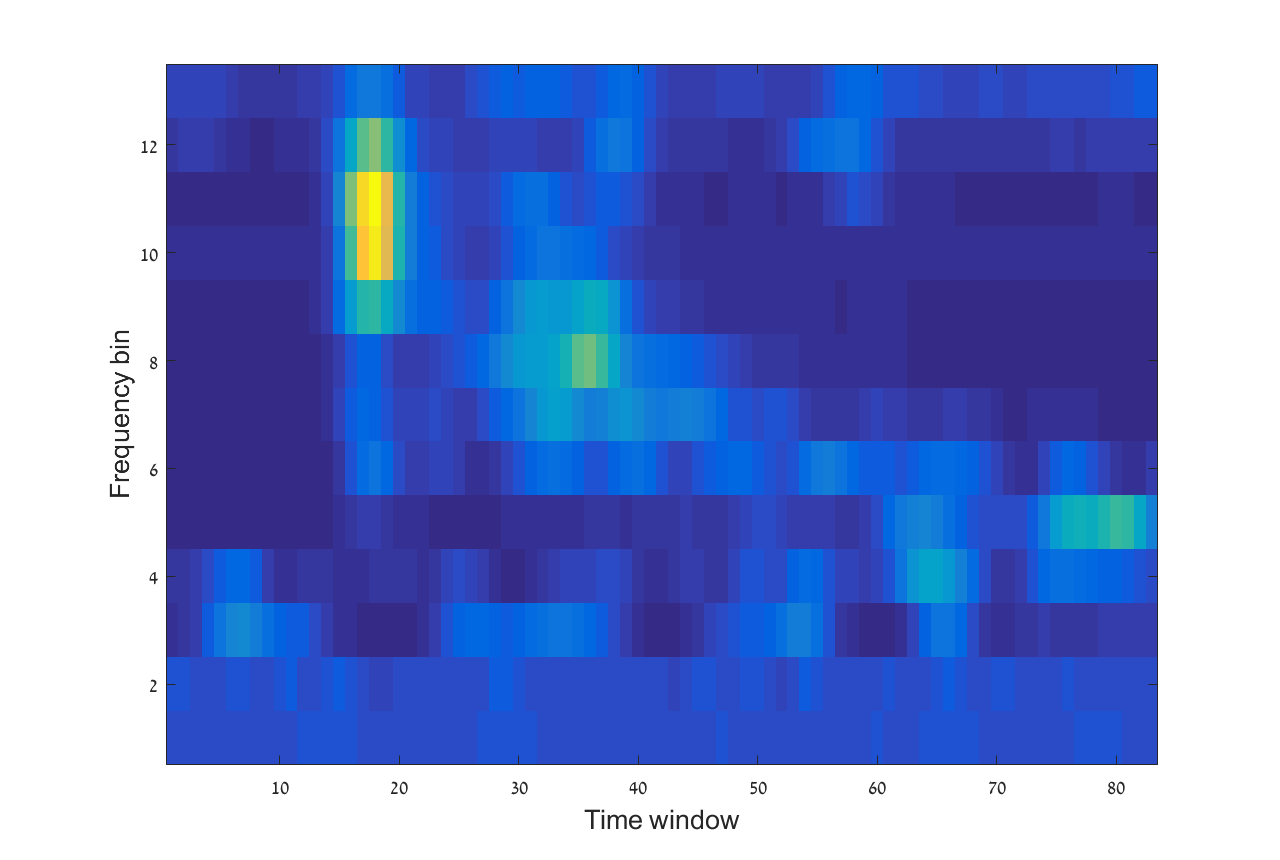}
	\caption{A sonogram extracted from the E channel of an explosion seismogram.}\label{fig:Sono}
\end{figure}
\begin{table}
	\centering
	
	\caption{The list of frequency bands used for the sonogram computation.}
	
	\label{table0}
	
	\begin{tabular}[t!]{|c|c|c|}
		\hline
		{ Band Number}           & f-start    & f-end    \\
		\hline
		{\#1}     & ${0}$ [Hz]	   & ${0}$	[Hz]	  \\
		{\#2}      & ${0.157}$	  [Hz]  & ${0.315}$	 [Hz]   \\
		{\#3}  	  	 & ${0.315}$	[Hz]    & ${0.630}$	    [Hz]  \\
		{\#4 }     & ${0.630}$	[Hz]   & ${1.102}$	[Hz]	  \\
		{\#5}      & ${1.102}$	[Hz]    & ${1.889}$	 [Hz] \\
		{\#6}  	   & ${1.889}$	[Hz]    & ${2.992}$	   [Hz]  \\
		{\#7}          & ${2.992}$	[Hz]    & ${4.567}$	  [Hz]    \\
		{\#8}          & ${4.567}$	 [Hz]   & ${6.772}$	 [Hz]   \\
		{\#9}          & ${6.772}$	 [Hz]    & ${9.921}$	[Hz]     \\
		{\#10}          & ${9.921}$	 [Hz]   & ${14.331}$	[Hz]      \\
		{\#11}          & ${14.331}$	[Hz]    & ${20}$	 [Hz]      \\
		\hline
	\end{tabular}
\end{table}
\item The bins are normalized such that the sum of energy in every frequency band is equal to $1$. The resulted sonogram is denoted by $\myvec{S}(k,t)$, where $k$ is the frequency band number, and $t$ is the time window number. Finally, we transpose the sonogram matrix into a Sonovector $\myvec{x}$ by concatenating the columns such that
\begin{equation} \label{eq:Sonovec}
\myvec{x}=\myvec{S}(:).
\end{equation}

} 
\end{itemize}
An example of a sonogram extracted from an explosion is presented in Figure \ref{fig:Sono}.

\section{Case Studies}
\label{sec:Exp}

To evaluate the strength of multi-view DM for identifying the properties of seismic events we perform the following experiments.
\subsection{Discrimination Between Earthquakes and Explosions}
\begin{figure}
	\centering
	\includegraphics[scale = 0.3]{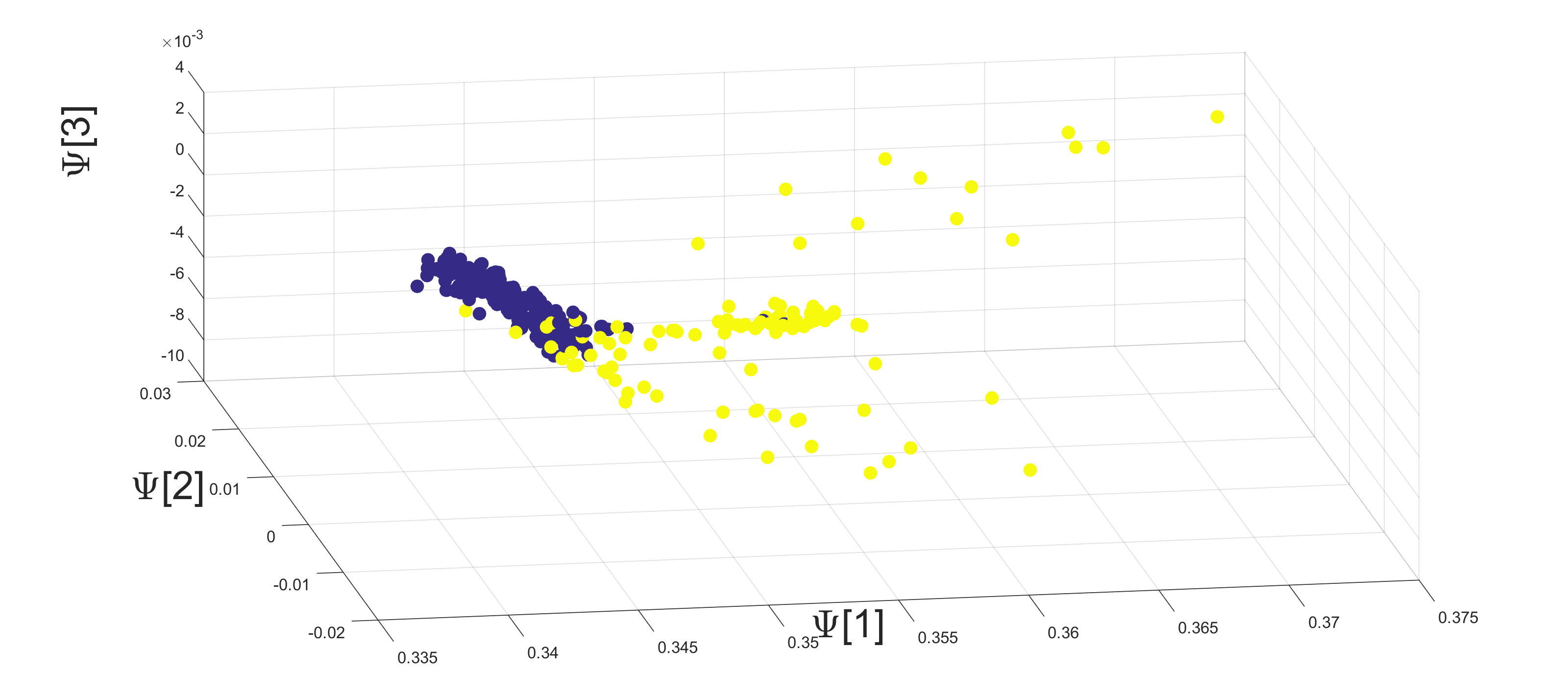}
	\caption{A 3-dimensional DM mapping extracted from recordings of the E channel. Blue points represent man-made explosions from a variety of sources. Yellow points represent recordings of earthquakes most of which were originated in southern part of Israel.}\label{fig:MapE}
\end{figure}

\begin{figure}
	\centering
	\includegraphics[scale = 0.2]{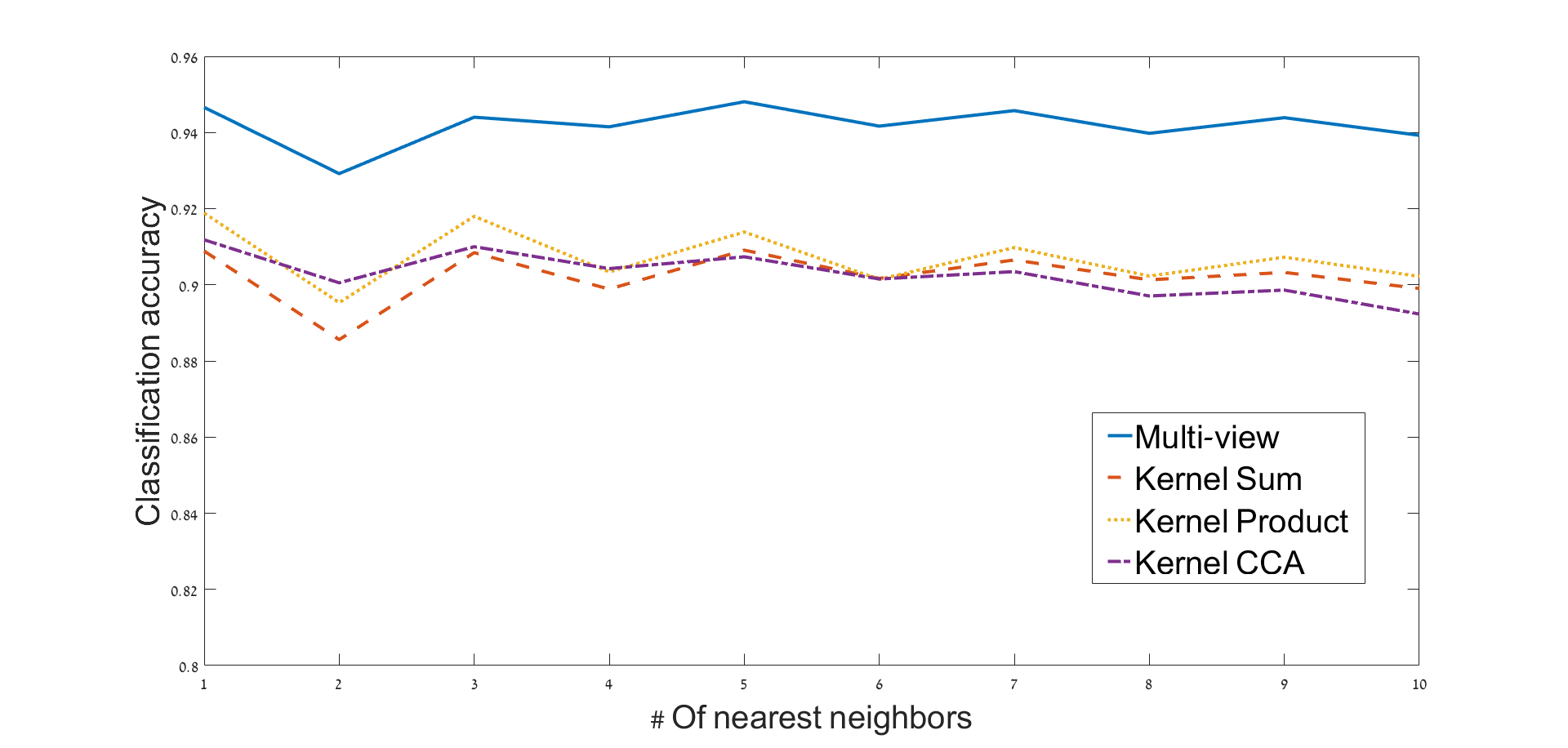}
	\caption{The classification accuracy for two classes, 105 earthquakes and 210 explosions.}\label{fig:ClassifyExpEq}
\end{figure}

We consider the earthquake-explosion discrimination problem as a supervised binary classification task. A homogeneous evaluation data set is constructed by using  data from 105 earthquakes and a random sample of 210 explosions. The sampling is repeated 200 times, and the results are the average of all trials. Algorithm \ref{alg:SeismicDM} is applied to extract a low dimensional representation of the seismic data. The number of data samples used for each events is 6000, where $N_1=1199$ (samples before onset) and $N_2=3800$ (samples after onset). An example of a 3-dimensional single view DM mapping is presented in Figure \ref{fig:MapE}. In this example, the explosions seem geometrically concentrated, while the earthquakes are spread out. This spread out structure may be associated with the diversity of the time-spectral information describing earthquakes, as oppose to the explosions that were mostly generated in specific quarries. The separation is clearly visible in this example. An evaluation of the separation is performed using a 1-fold cross-validation procedure.  Test points are classified by using a simple K-NN classifier in a $d=4$ dimensional representation. The optimal dimension ($d=4$) for classification was found empirically based on our data set. The average accuracy of classification for various values of $K$ are presented in Figure \ref{fig:ClassifyExpEq}.  Thus, the multi-view approach shows better performance with 95\% of correct discrimination. 

\begin{algorithm}
	\caption{Mapping of seismic data} \textbf{Input:} Three sets of time signals $\myvec{Y}_E, \myvec{Y}_N,\myvec{Y}_Z$. One for each seismic channel.\\ \textbf{Output:} A low dimensional mapping $\myvec{\Psi}(\myvec{Y}_E,\myvec{Y}_N,\myvec{Y}_Z)$.
	\begin{algorithmic}[1]
		\STATE Apply Algorithm \ref{alg:Trigger} to each time signal $\myvec{y}_Z^{(i)}$ and estimate the P onset $\hat{n}^{(i)}$.
		\STATE Define the aligned truncated signal as $\myvec{\bar{y}}_Z^{(i)}(n)\defeq[y_Z^{(i)}(\hat{n}^{(i)}-N_1),...,y_Z^{(i)}(\hat{n}^{(i)}+N_2)]$.
        \STATE Compute $\myvec{\bar{y}}_E^{(i)}(n) \text{ and } \myvec{\bar{y}}_N^{(i)}(n)$ in a similar manner.
        \STATE Compute the Sonovecs  based on Eqs. (\ref{eq:STFT}), (\ref{eq:NSPEC}) and (\ref{eq:Sonovec}).
        \STATE Compute the DM mappings $\myvec{\Psi}_E,\myvec{\Psi}_N,\myvec{\Psi}_Z$ (Eq. (\ref{EQPSI})).
        \STATE Compute the multi-view DM mapping $\vec {\myvec{\Psi}} $ (Eq. (\ref{eq:MVDMrep})).
	\end{algorithmic}
	\label{alg:SeismicDM}
\end{algorithm}

\subsection{Quarry Classification}
\begin{figure}
	\centering
	\includegraphics[scale = 0.4]{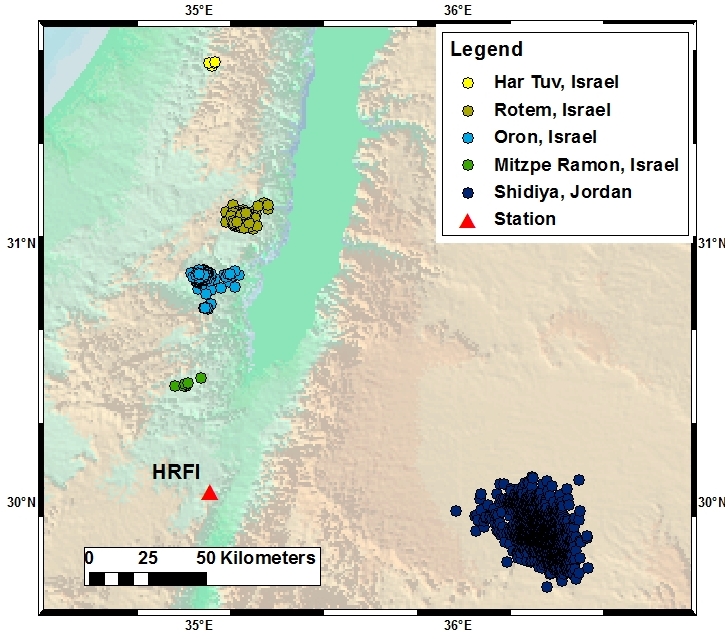}
	\caption{Map of quarry clusters.}\label{fig:QuarryLocations}
\end{figure}

Identification and separation of quarries by attributing the explosions to the known sources is a challenging task in  observational seismology \cite{Harris1}, \cite{Harris2}. Here we demonstrate how  the DM representation can be utilized to identify the origin of an explosion.
\begin{table} \label{Table:Loc}\caption{Description of quarry clusters.}
	\begin{tabular}[h!]{|l|c|c|c|c|}
	\hline
	{Quarry Clusters}           & \# of events   & Center Lat  &  Center Lon  & Distances to HRFI \\
	\hline
	{Shidiya, Jordan }     & ${250}$	   & ${29.91}^{\circ}$			& ${36.32}^{\circ}$  & $125\text{[Km]}$   \\
	{Oron, Israel }      & ${222}$	    & ${30.82}^{\circ}$	     & ${35.04}^{\circ}$ & $86.7\text{[Km]}$  \\
	{Rotem, Israel }  	  	& ${115}$	    & ${31.09}^{\circ}$	     & ${35.19}^{\circ}$  & $117.7\text{[Km]}$  \\
   	{M. Ramon, Israel }  	  	 & ${8}$	    & ${30.46}^{\circ}$	     & ${34.95}^{\circ}$  & $47.3\text{[Km]}$  \\
    {Har Tuv, Israel }  	  	 & ${7}$	    & ${31.68}^{\circ}$	     & ${35.05}^{\circ}$  & $128.2\text{[Km]}$  \\

	\hline
\end{tabular}
\end{table}
For this study  602 seismograms  of explosions are used. The explosions occurred in 5 quarry clusters (see Table \ref{Table:Loc} and Figure \ref{fig:QuarryLocations}) and the label data was taken from seismic catalogs. It should be noted that the quarry clusters may include several neighboring quarries and the quarry area may be of several kilometers (like Rotem) or more than ten kilometers (like Shidiya).  Moreover, the precise (“ground truth”) location for most of explosions inside a quarry are not known. We estimate that the hypo-center accuracy in the used seismic catalogs is about a few kilometers for the explosions in Israel and it is more than ten kilometers for the explosions in Jordan, which are located outside the Israeli seismic network. The mean latitude and longitude are computed  for the explosions belonging to each cluster and referred them to the nearby quarry (see Table \ref{Table:Loc}). 

\begin{figure}
	\centering
	\includegraphics[scale = 0.24]{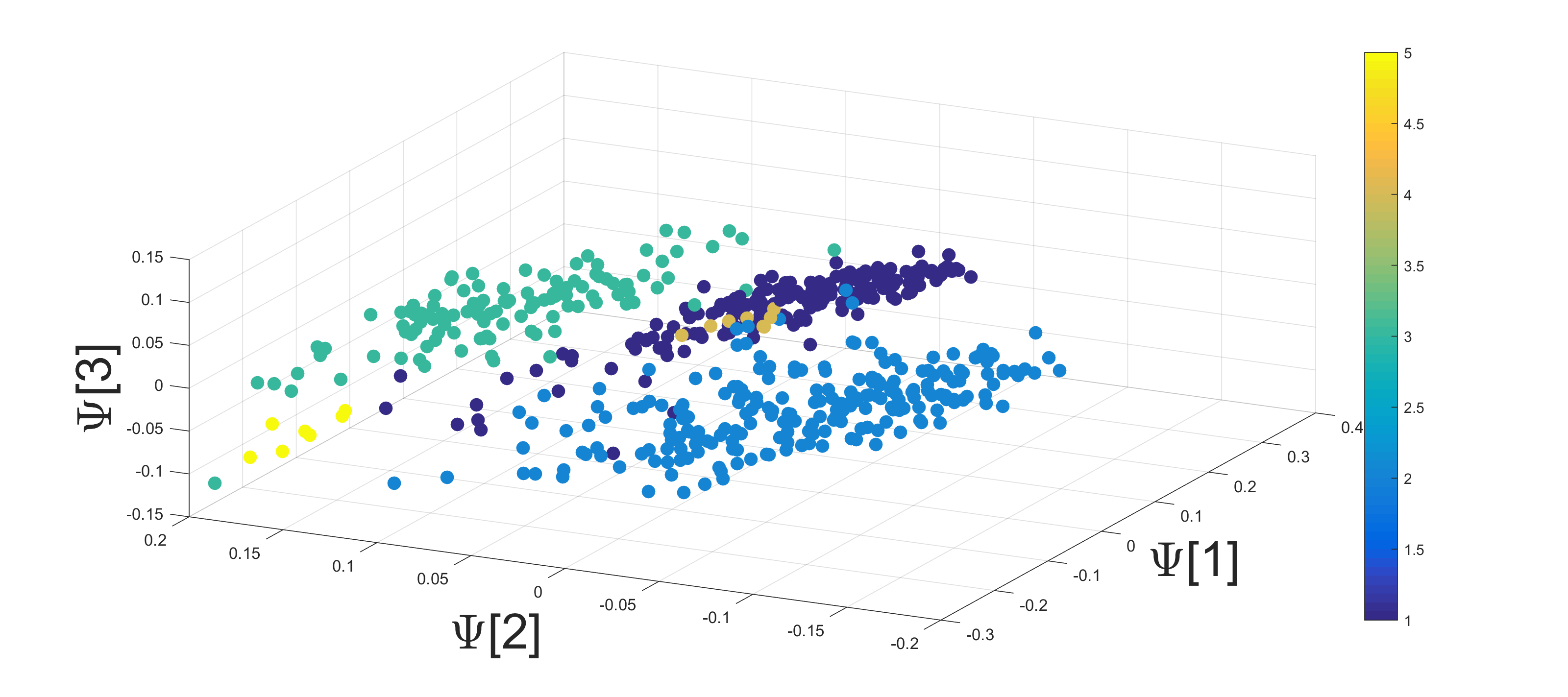}
	\caption{A 3-dimensional diffusion mapping of 602 explosions.}\label{fig:QuarriesMapZ}
\end{figure}

The application of Algorithm \ref{alg:SeismicDM} yields a low dimensional representation of the seismic recordings. An example of a 3-dimensional single view DM mapping is presented in Figure \ref{fig:QuarriesMapZ}.

\begin{figure}
	\centering
	\includegraphics[scale = 0.25]{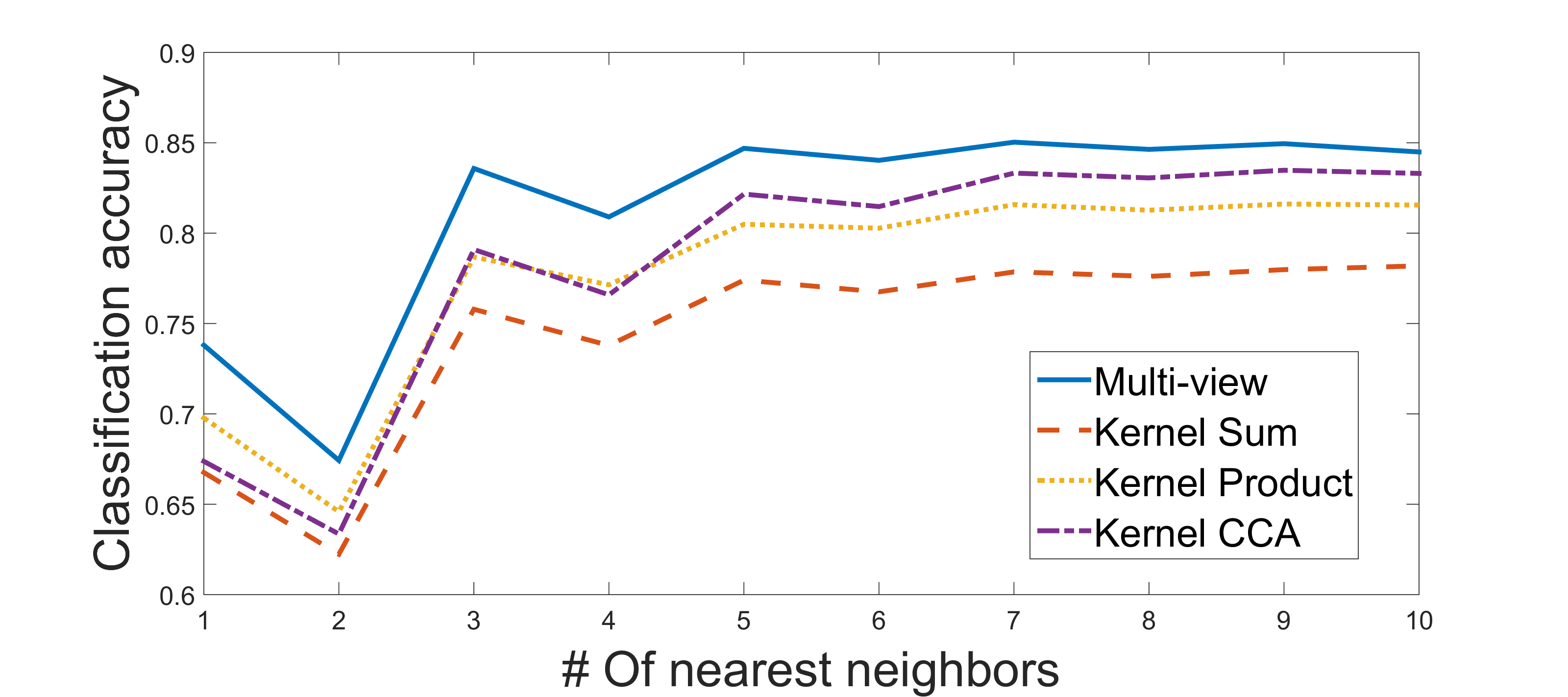}
	\caption{The classification accuracy for 5 source locations.}\label{fig:ClassifyQuarry}
\end{figure}

The mapping is followed by a classification step that is performed based on a 1-fold cross validation using K-NN with $K=3$. The accuracy of the classification is presented in Figure \ref{fig:ClassifyQuarry}. The multi-view approach shows a peak performance of 85\% of correct classification rate. 

\subsection{Location Estimation}
The following case study demonstrates how the diffusion coordinates extract underlying physical properties of the sampled signal. In particular we show that the low dimensional representation that is generated by diffusion maps organizes the events with respect to their source location, even though this was not an input parameter of the algorithm. The original high-dimensional space holds the sonogram of each event.  Nearly co-located events with the similar source mechanisms and magnitudes should have a similar time-frequency content and, consequently, have similar sonograms. Therefore, we expect them to lie close to each other in the high dimensional space. The diffusion distance, which is the metric that is preserved in DM, embeds the data while keeping its geometrical structure. Thus, physical properties  (such as the source location) that characterize the sonogram and therefore define the geometric structure of the points in the high-dimensional space, are preserved in low-dimensional DM embedding. Note that such a geometry preserving metric does not exist in linear dimensionality reduction methods like PCA. 

The dataset for this study includes 352 explosions that occurred in 4 quarry clustering Israel out 5 clusters above. The explosions in Jordan were removed since they are located at a large distance from the HRFI station. We show that the location of seismic events can be evaluated from the DM embedding coordinates. A similar evaluation based on a linear projection that was calculated with PCA yields a less accurate correlation to the events' true location.   

\begin{figure}

\includegraphics[width=3.7in]{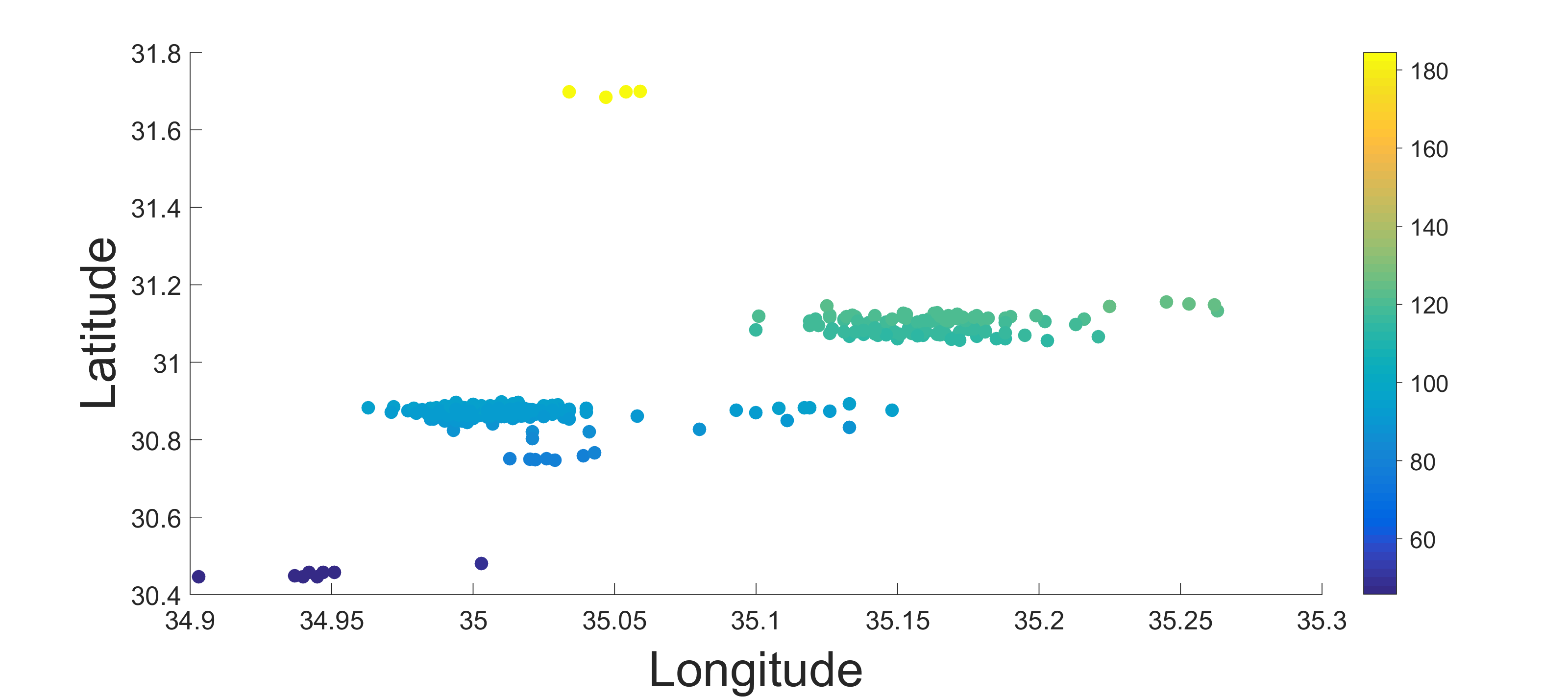}
\includegraphics[width=3.7in]{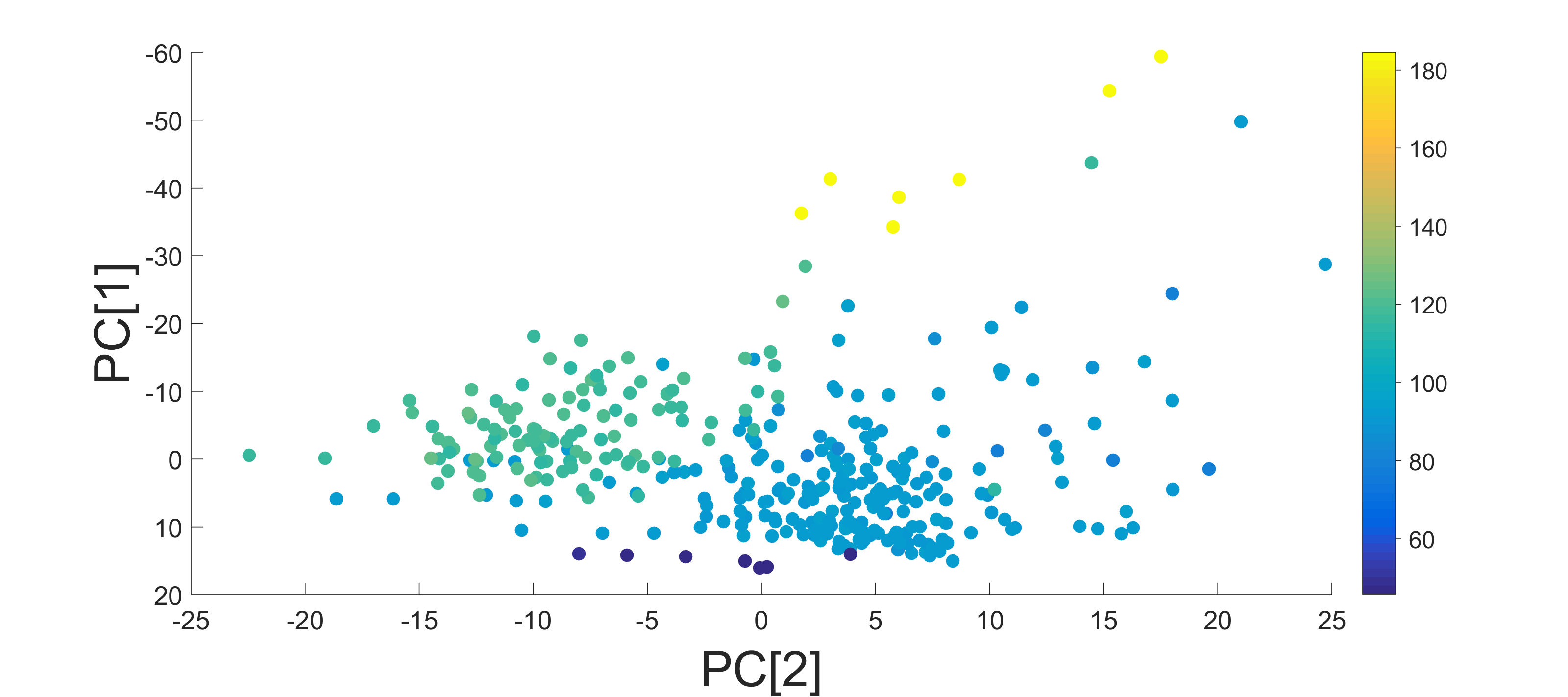}
\includegraphics[width=3.7in]{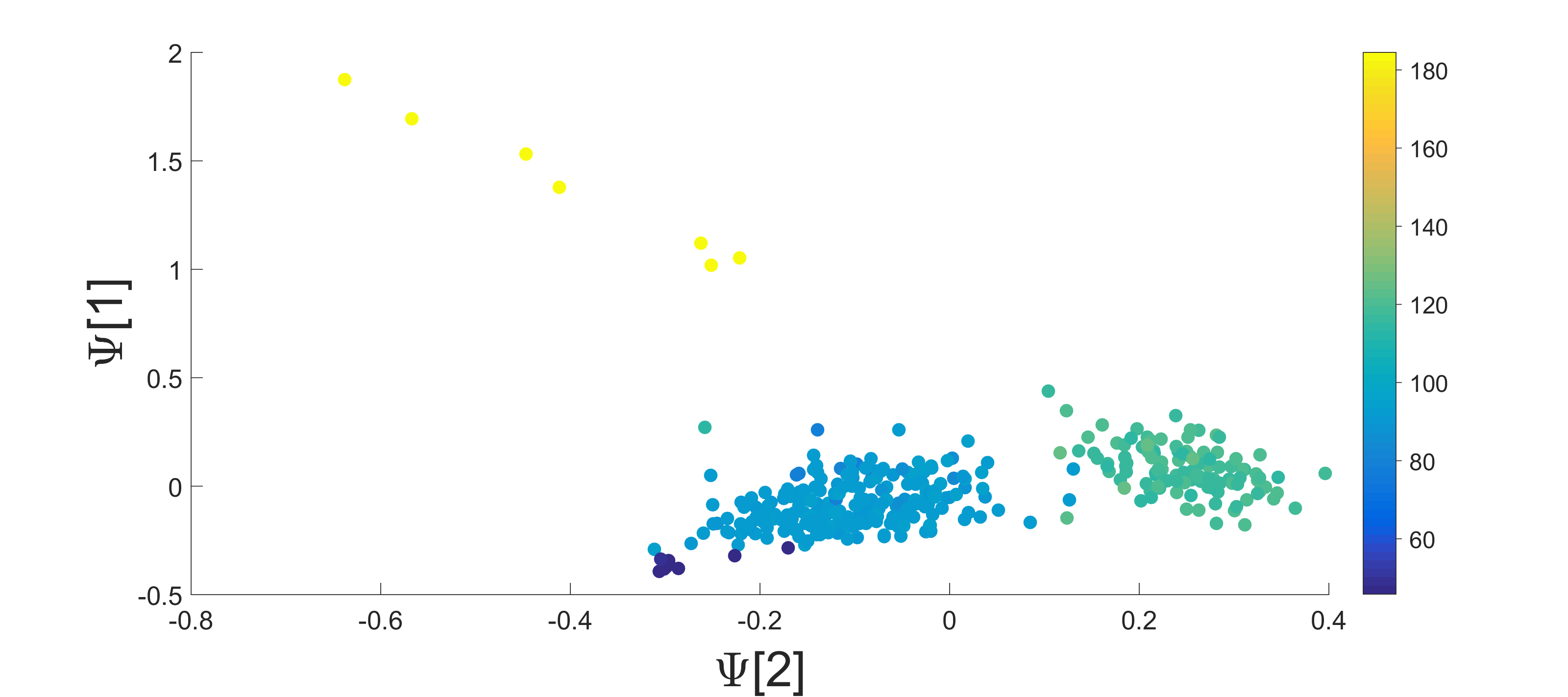}
\caption{Top- the manually estimated location of events. Middle- the first two principle components  of the N-channel. Bottom- the first two diffusion coordinated of the N-channel. Color represents the distance from HRFI station.}
\label{fig:MapLoc}
\end{figure}

Figure \ref{fig:MapLoc} (top image) displays the longitude and latitude coordinates of catalog locations of the events. These are the source locations of the seismic events. The points are colored by distance in kilometers from HRFI station. The middle and bottom images of Figure \ref{fig:MapLoc} present the two-dimensional PCA and DM embeddings of the dataset, respectively. It is clearly evident that the DM (bottom image in Figure \ref{fig:MapLoc}) representation has captured the location variability, while in the PCA representation this intrinsic factor is less obvious (middle image in Figure \ref{fig:MapLoc}). In the DM embedding, the clusters are well separated with respect to the event's location. In PCA the separation is not as clear, meaning that the low-dimensional PCA representation does not reveal this property.  The Pearson correlation coefficients between first two diffusion coordinates and relative latitude and longitude are 0.82 and 0.77 for both dimensions respectively. The Pearson correlation coefficients between first two principle and relative latitude and longitude are 0.56 and 0.39 respectively.

\subsection{Detecting Anomalous Events}
This case study demonstrates the diffusion representation's ability to detect anomalous events among set of events at specific site. When two events are nearly co-located, have close magnitudes but with different source mechanisms, then their sonograms should be quite different as well.

Ripple-fire explosions are part of routine mining production cycles at the Oron phosphate quarry in Israel. In July 2006, three experimental one shot explosions were conducted by the Geophysical Institute of Israel at the Oron quarry \cite{gitterman2009source}.  Our goal is to distinguish between the one shot explosions and the ripple-fire quarry blasts. This is not a trivial task, as all the events were conducted at very close distances. 

\begin{figure}
	\centering
    \includegraphics[scale = 0.25]{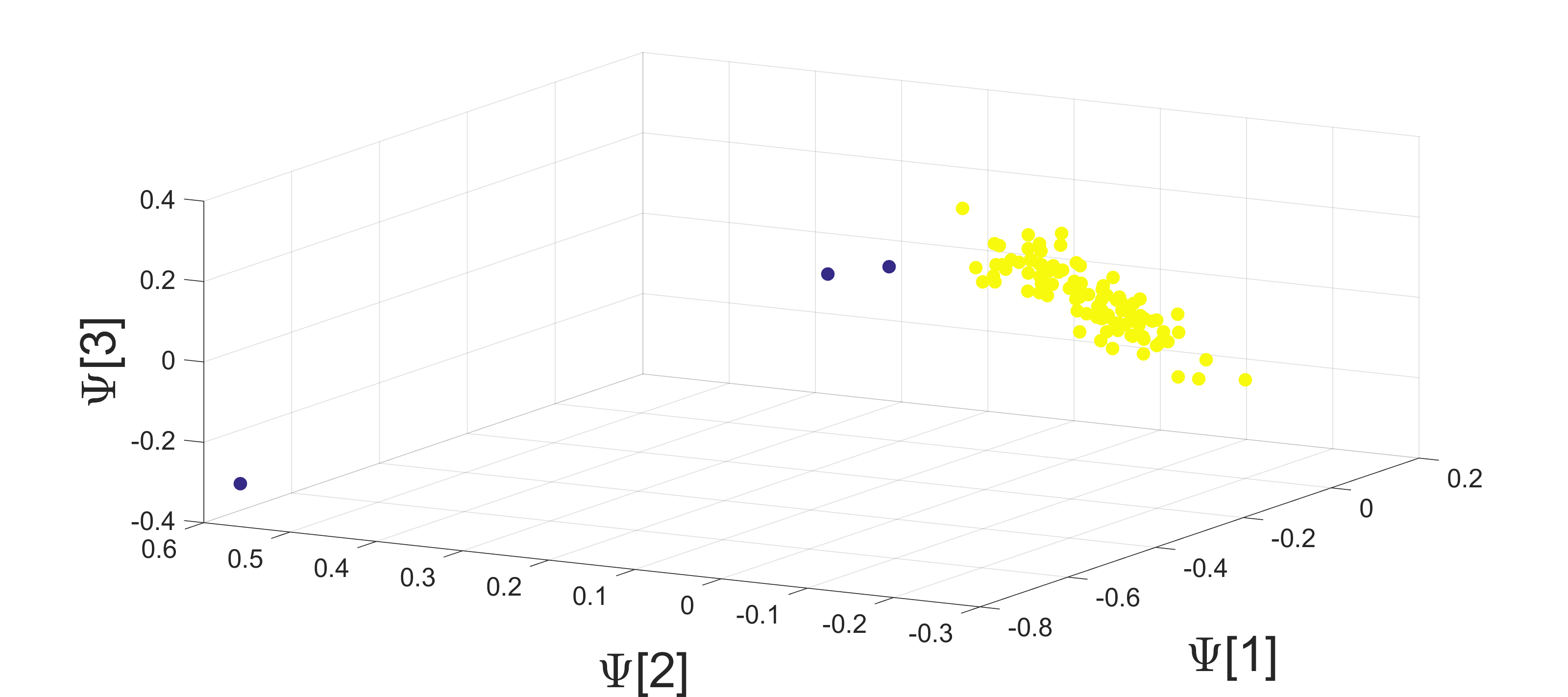}
	\caption{Diffusion representation of 98 explosions recorded using the Z-channel. The suspected anomalies are colored in blue.}\label{fig:Anomalies}
\end{figure}

\begin{figure}
	\centering
	\includegraphics[scale = 1]{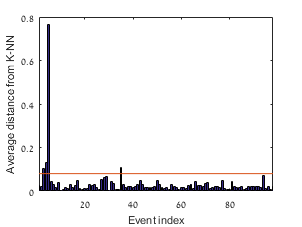}

	\caption{Average K-NN distance for each explosion. The distance is computed using $d=3$ coordinates and $K=5$ nearest neighbors.}\label{fig:AverageKNN}
\end{figure}

\begin{algorithm}
	\caption{K-NN based anomaly detection} \textbf{Input:} Low dimensional mapping $\myvec{\Psi}$.\\ \textbf{Output:} A set of indexes $\cal{I}$ of suspected anomalies.
	\begin{algorithmic}[1]
		\STATE Find $K$ nearest neighbors for all data points $\myvec{\bar{\Psi}}(\myvec{y}_i),i=1,...,M$, denote the set as $\cal{J}$.
		\STATE Define the K-NN average distance as $\myvec{\hat{D}}_i\defeq \sum^K_{l=2} \frac{||\myvec{\Psi}(\myvec{y}_i)-\myvec{\Psi}(\myvec{y}_{j_l})||^2}{K} $.
        \STATE Find all points with average distance $\myvec{\hat{D}}_i$ larger then a threshold $\delta$.

	\end{algorithmic}
	\label{alg:KNNanomaly}
\end{algorithm}
To remove the variability created by the location of the events, 98 blast from a small region surrounding the ground truth location of the experimental explosions as reported in \cite{gitterman2009source} are used. Algorithm \ref{alg:SeismicDM} is applied and a mapping extracted from the Z-channel is used to find the suspected anomalies. The diffusion maps embedding is presented in Figure \ref{fig:Anomalies}. The three anomaly points are colored in blue, they are clearly separated from the main cluster. The anomalies are automatically identified using Algorithm \ref{alg:KNNanomaly} with $K=4$ and a threshold set as four times the median of all distances $\hat{D}_i,i=1,...,M$. The average K-NN distance for the 98 blasts is presented in Figure \ref{fig:AverageKNN}. The four events that were suspected as anomalies include the three experimental explosions (which are described in \cite{gitterman2009source}).

\section{Conclusion}
\label{sec:Future}
In this paper, we have adapted a multi-view manifold learning framework for fusion of seismic recordings and for low-dimensional modeling. The abilities of kernel fusion methods for extracting meaningful seismic parameters were demonstrated on various case studies. Various algorithms for classification of seismic events type, location estimation and anomaly detection were presented. These algorithms can be used as decision support tools for analysts who need to determine the source, location and type of recorded seismic events. Correct classification of events results in improved and more accurate seismic bulletins.

The proposed method is model free, thus it does not require knowledge of physical parameters. The underlying physical parameters are revealed by the diffusion maps and multi-view constructions. This type of kernel based sensor fusion is new in seismic signal processing and it overcomes some of the limitation of traditional model based fusion methods. 
\section*{Acknowledgments}
	This research was supported by the research grant of Pazy Foundation. We would like to thank Yochai Ben Horin for his advice and suggestions. We are grateful to Dov Zakosky and Batia Reich for providing us with the seismic catalog of Geophysical Institute of Israel.

\ifCLASSOPTIONcaptionsoff
  \newpage
\fi



\end{document}